\documentclass{article}
\usepackage{amsmath, amssymb}
\usepackage{graphicx}
\usepackage{hyperref}
\usepackage{float}
\usepackage{authblk}
\usepackage{caption}
\usepackage{subcaption}
\usepackage{placeins}

\title{LLM Generated Distribution-Based Prediction of \\ US Electoral Results, Part I}

\author{Caleb Bradshaw}
\author{Caelen Miller}
\author{Sean Warnick}

\affil{Brigham Young University, Provo, Utah}

\date{November 4, 2024}

\begin{document}
\maketitle

\begin{abstract}
This paper introduces \textit{distribution-based prediction}, a novel approach to using Large Language Models (LLMs) as predictive tools by interpreting output token probabilities as distributions representing the models' learned representation of the world. This distribution-based nature offers an alternative perspective for analyzing algorithmic fidelity, complementing the approach used in silicon sampling.
We demonstrate the use of distribution-based prediction in the context of recent United States presidential election, showing that this method can be used to determine task specific bias, prompt noise, and algorithmic fidelity. This approach has significant implications for assessing the reliability and increasing transparency of LLM-based predictions across various domains.
\end{abstract}

\section{Introduction}
Language is a symbolic system used to convey semantics—a vehicle for representing meaning. By translating thoughts, experiences, and concepts into structured expressions, language serves as a bridge between abstract ideas and tangible communication. Large Language Models (LLMs) extend this function into the realm of artificial intelligence, demonstrating emergent abilities to handle complex and context rich tasks beyond the scope of smaller language models through natural language \cite{wei2022emergentabilitieslargelanguage}. Furthermore, research indicates that LLMs contain coherent and grounded representations that reflect real world distributions \cite{gurnee2024languagemodelsrepresentspace, li2024emergentworldrepresentationsexploring, borisov2023languagemodelsrealistictabular}. This emergent property of LLMs is demonstrated in Appendix~\ref{appendix:model_size}. These internal world models are informed strongly by training data, frequently resulting in significant biases \cite{feng2023pretrainingdatalanguagemodels, rozado2024politicalpreferencesllms, RePEc:osf:socarx:97r8s}. The accurate representation of real world distributions through LLM outputs is known as algorithmic fidelity, a concept that has fueled methods such as silicon sampling, the generation of simulated personas in an attempt to simulate real populations \cite{Argyle_2023}. This technique leverages the internal model of LLMs to simulate human-like responses, potentially providing insights into diverse demographic samples without the need for actual data. 

A key limitation of silicon sampling lies in its tendency to stereotype demographics, as demonstrated in Appendix~\ref{appendix:silicon_sampling}. Additionally, recent results have suggested that the success of silicon sampling is due in part to input shortcut features \cite{yang2024largelanguagemodelsllms}. Prompt noise, such as the ordering of options, can also significantly impact voting results \cite{yang2024llmvotinghumanchoices, sun2024investigationpromptvariationszeroshot}. Furthermore, LLMs tend to assume that people act more rationally then they actually do, thereby deviating from actual human behavior \cite{liu2024largelanguagemodelsassume}. Despite these challenges, silicon sampling has powerful use cases, and when properly applied can generate realistic survey results \cite{sun2024randomsiliconsamplingsimulating, marketingsilsample2024, ong2024gptologycomputationalmodelssilicon}.

Building on these insights, our paper introduces a novel approach that leverages LLMs as non-persona based predictive models. Rather than having the model simulate individuals, we analyze output probabilities as predictive distributions, interpreting them as indicative of model expectations. We term this approach Distribution Based Prediction. This perspective enables a unique set of investigations, including bias detection through distributional analysis, examination of case by case algorithmic fidelity, robustness testing against variations in prompt design, and evaluation of LLMs’ effectiveness as predictive tools. 

To demonstrate this approach, we apply it to a real-world prediction task: predicting the outcome of the U.S. presidential election by generating distributions of voter share per state for each candidate. Voter prediction is a significant task for testing the efficacy of LLMs at gathering, synthesizing, and making predictions with data \cite{gujral2024llmshelppredictelections, RePEc:osf:socarx:97r8s}. This context serves as both a testing ground for an LLMs predictive accuracy and a framework to analyze bias, prompt noise sensitivity, and algorithmic fidelity.

\section{Methods}

\subsection{Distribution Based Prediction}

Given the limitations observed in demographic-based sampling, we propose a distribution based prediction methodology that bypasses individual persona simulation. Instead, we prompt the model directly to forecast electoral outcomes by requesting specific vote shares for each candidate within a designated state, treating the output token probabilities as a distribution representative of the model's embedded knowledge of demographic, geographic, and historical voting trends.

In this approach, we formulate a straightforward, task-oriented prompt structure aimed at obtaining a vote distribution across regions while allowing the model to express uncertainty implicitly in its probability outputs. Specifically, we provide the model with a “system prompt” role as a neutral election predictor, with a targeted output representing candidate vote share as a single integer token:

\begin{itemize}
\item \textbf{System Prompt:} \\
\textit{``You are an impartial election prediction machine. Respond with a single integer token between 0 and 100 representing the vote share."}

\item \textbf{User Prompt:} \\
\textit{``{candidate} is running against {opponent}, what percentage of the vote will \{candidate\} win in the \{year\} presidential election in \{state\}?"}
\end{itemize}

This distribution based prediction method is designed to leverage the model as a predictive agent rather than as an individual demographic persona. By eliciting responses in a probabilistic format, the model is encouraged to reflect its inherent understanding of regional electoral dynamics and demographic tendencies. Empirically, this method produces plausible distributions that change for different geographies, demographics, time periods, and candidate choices.

\FloatBarrier
\subsection{Condensing State Distributions}
Let \( S = \{s_1, s_2, \ldots, s_{51}\} \) denote the set of states, including the District of Columbia, and let \( C = \{c_1, c_2\} \) represent the two candidates. Define the discrete voter percentage set as \( Y = \{0\%, 1\%, 2\%, \ldots, 100\%\} \).

For each state \( s \in S \) and candidate \( c \in C \), let \( P_{s,c}(y) \) denote the probability that candidate \( c \) receives exactly \( y\% \) of the vote in state \( s \). These probability distributions satisfy the normalization condition:
\begin{equation*}
\sum_{y \in Y} P_{s,c}(y) = 1 \quad \forall s \in S, \, c \in C
\end{equation*}

The objective is to compute the win probabilities for each candidate in each state by determining the likelihood that one candidate's voter percentage exceeds the other's. Tied outcomes are excluded from consideration.

The probability that candidate \( c_1 \) wins in state \( s \) is given by:
\begin{equation*}
P_{s}(c_1 \text{ wins}) = \sum_{y_1 \in Y} P_{s,c_1}(y_1) \cdot \left( \sum_{y_2 < y_1} P_{s,c_2}(y_2) \right)
\end{equation*}

Similarly, the probability that candidate \( c_2 \) wins in state \( s \) is:
\begin{equation*}
P_{s}(c_2 \text{ wins}) = \sum_{y_2 \in Y} P_{s,c_2}(y_2) \cdot \left( \sum_{y_1 < y_2} P_{s,c_1}(y_1) \right)
\end{equation*}

Since ties are disregarded and only one candidate can win in each state, the win probabilities satisfy:
\begin{equation*}
P_{s}(c_1 \text{ wins}) + P_{s}(c_2 \text{ wins}) = 1 \quad \forall s \in S
\end{equation*}

For each state \( s \in S \), the win probabilities for candidates \( c_1 \) and \( c_2 \) are computed as:
\begin{align*}
P_{s}(c_1 \text{ wins}) &= \sum_{y_1 \in Y} P_{s,c_1}(y_1) \cdot \left( \sum_{\substack{y_2 \in Y \\ y_2 < y_1}} P_{s,c_2}(y_2) \right), \\
P_{s}(c_2 \text{ wins}) &= \sum_{y_2 \in Y} P_{s,c_2}(y_2) \cdot \left( \sum_{\substack{y_1 \in Y \\ y_1 < y_2}} P_{s,c_1}(y_1) \right).
\end{align*}

These equations ensure that for each state \( s \), the sum of the win probabilities for both candidates equals unity, thereby providing a complete probabilistic model of the election outcomes based on discrete voter percentages.

\subsection{Electoral College Simulation}
Let \( S = \{s_1, s_2, \ldots, s_{51}\} \) denote the set of states, including the District of Columbia, where each state \( s \in S \) is assigned a fixed number of electoral votes \( e_s \). Define the total number of electoral votes as \( E = \sum_{s \in S} e_s \).

For each state \( s \), let \( P_s(c_1) \) and \( P_s(c_2) = 1 - P_s(c_1) \) represent the probabilities that candidates \( c_1 \) and \( c_2 \) win state \( s \), respectively. Define a random variable \( V_s \) to represent the electoral votes allocated to \( c_1 \) from state \( s \):
\[
V_s =
\begin{cases}
e_s & \text{with probability } P_s(c_1), \\
0 & \text{with probability } P_s(c_2).
\end{cases}
\]
The total electoral votes \( V \) for candidate \( c_1 \) is the sum of electoral votes from all states:
\[
V = \sum_{s \in S} V_s.
\]
The probability that \( c_1 \) receives exactly \( k \) electoral votes is given by the probability mass function:
\[
P(V = k) = \sum_{\substack{A \subseteq S \\ \sum_{s \in A} e_s = k}} \left( \prod_{s \in A} P_s(c_1) \right) \left( \prod_{s \in S \setminus A} P_s(c_2) \right).
\]
Due to the computational complexity of enumerating all subsets \( A \subseteq S \), a generating function approach is employed. The generating function \( G_V(z) \) for \( V \) is defined as:
\[
G_V(z) = \prod_{s \in S} \left[ P_s(c_2) + P_s(c_1) \cdot z^{e_s} \right].
\]
The probability \( P(V = k) \) is the coefficient of \( z^k \) in the expanded form of \( G_V(z) \):
\[
P(V = k) = [z^k] G_V(z).
\]
For example, if there is a \( 5\% \) probability that \( c_1 \) accumulates \( 272 \) electoral votes, then \( c_2 \) has a \( 5\% \) probability of receiving \( E - 272 \) electoral votes:
\[
P(V = 272) = 0.05 \quad \Rightarrow \quad P(V = E - 272) = 0.05.
\]
Finally, the probabilities across all possible electoral vote outcomes satisfy the normalization condition:
\[
\sum_{k=0}^{E} P(V = k) = 1.
\]
This formulation provides a comprehensive framework for determining the probability distribution of Electoral College outcomes based on state-level win probabilities.

\section{Experiments and Results}
\subsection{Comparison to 2020 Election Results}

Our model's performance in "predicting" the 2020 U.S. presidential election provides a useful benchmark for evaluating its algorithmic fidelity on historic data. When prompted to forecast state-by-state outcomes for the candidates, the model generated distributions that were notably concentrated around the actual vote percentages, with most states showing over 90\% of their probability mass within a percent of the true results.

\begin{figure}[H]
    \centering
    \includegraphics[width=0.6\textwidth]{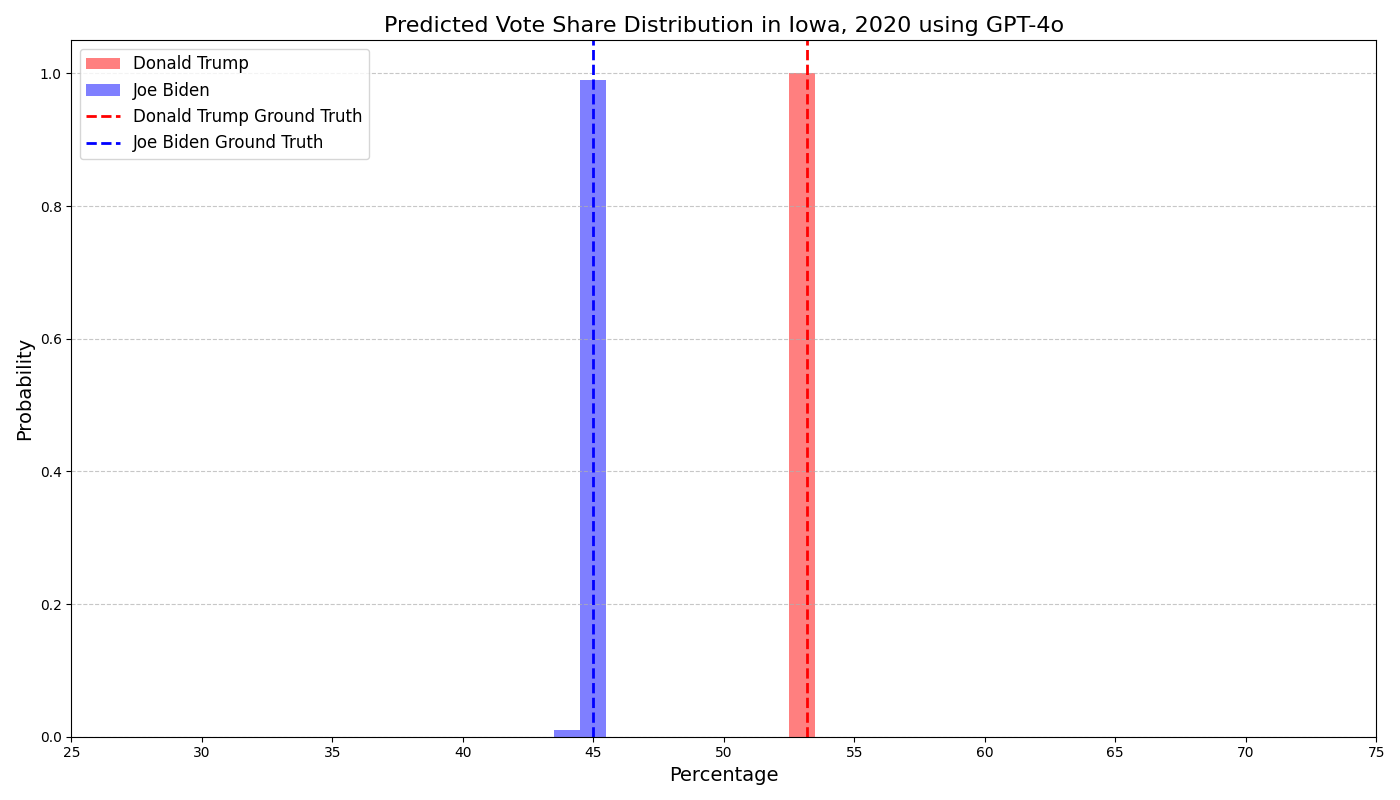}
    \caption{Model predictions for Iowa in the 2020 presidential election, showing a high concentration of probability mass around the actual vote percentage.}
    \label{fig:iowa_accuracy}
\end{figure}

We analyze the distributional error by comparing the model’s predicted vote shares for Joe Biden and Donald Trump with the actual percentages from 2020. As shown in Table~\ref{tab:state_differences}, the model achieved an average prediction error of 0.4490 for Joe Biden and 0.4354 for Donald Trump across all states, meaning that the model's outputs deviate by less than half a percentage point on average. This precision and accuracy is evident in Iowa (see Figure~\ref{fig:iowa_accuracy}), where the predicted distributions tightly align with the actual vote shares.

\begin{figure}[H]
    \centering
    \begin{minipage}{0.45\textwidth}
        \centering
        \includegraphics[width=\textwidth]{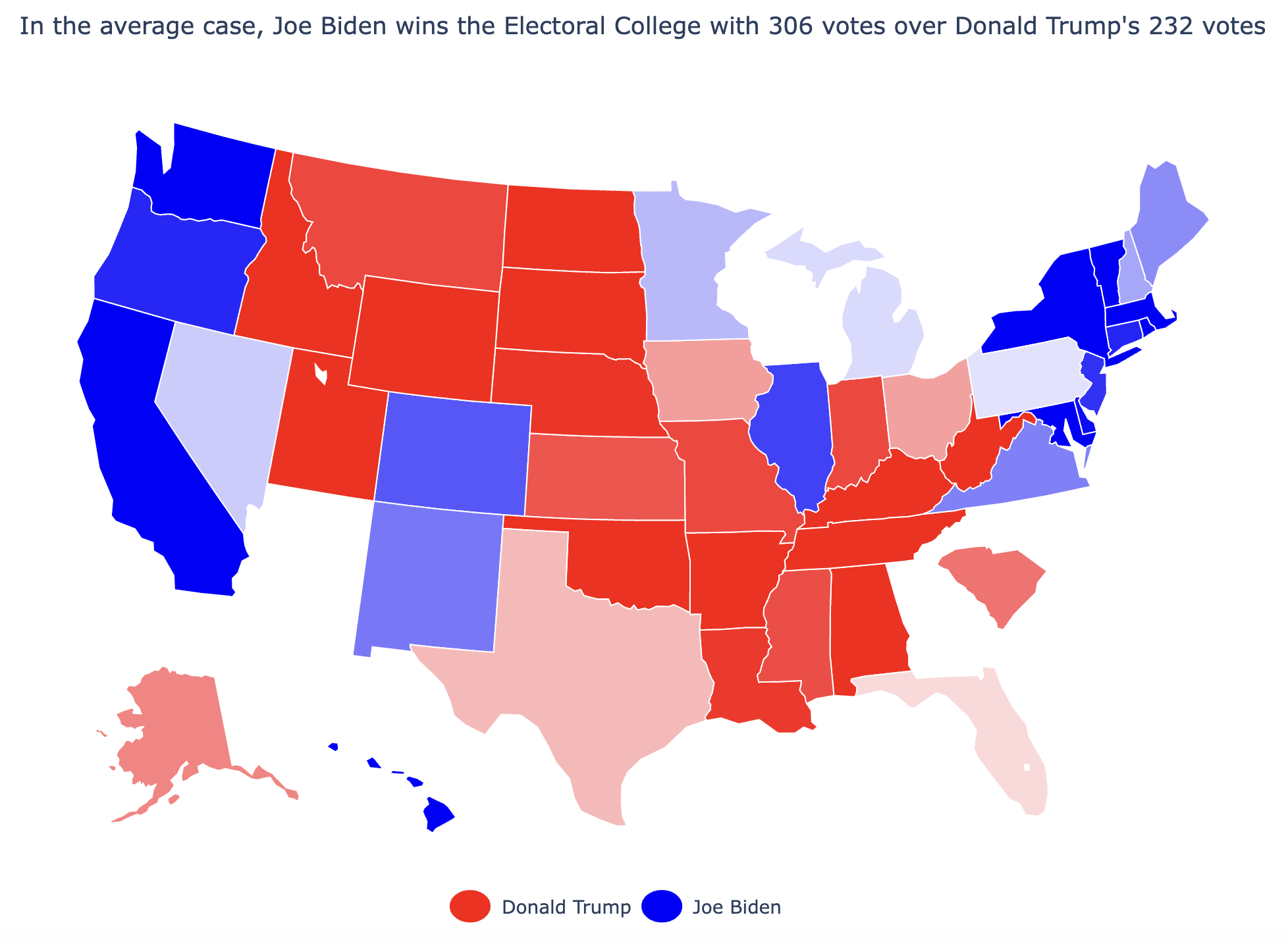}
        \caption{Model-predicted electoral college outcome for the 2020 election, matching the actual results.}
        \label{fig:model_prediction_2020}
    \end{minipage}\hfill
    \begin{minipage}{0.45\textwidth}
        \centering
        \includegraphics[width=\textwidth]{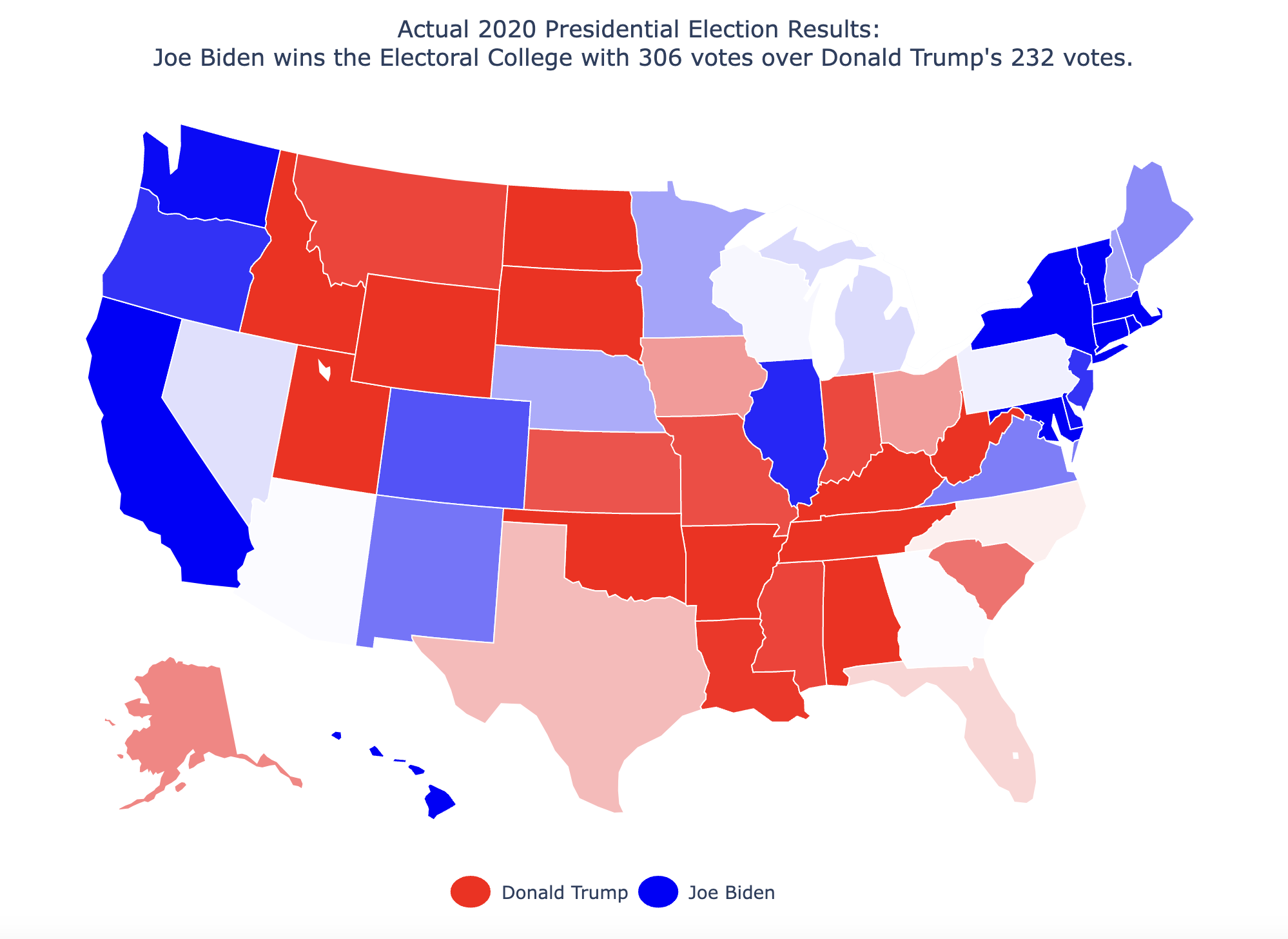}
        \caption{Actual electoral college outcome for the 2020 election.}
        \label{fig:ground_truth_2020}
    \end{minipage}
\end{figure}

As demonstrated in Figure~\ref{fig:model_prediction_2020} and Figure~\ref{fig:ground_truth_2020}, the model's predicted electoral college outcome for the 2020 election aligns with the actual results. This accuracy underscores the model's capacity to reflect real-world voting behavior correctly when aggregating predictions at the state level. Such alignment highlights the robustness of the LLM's internal world model in capturing the nuances of the electoral process.

At least in the context of historical state level voting data for the presidential election, the model demonstrates a significant degree of algorithmic fidelity.

\subsection{State Distributions for the 2024 Election}
We now prompt the model on the future 2024 election between Donald Trump and Kamala Harris to get expected probability distributions for each state.

\begin{figure}[H]
    \centering
    \includegraphics[width=0.6\textwidth]{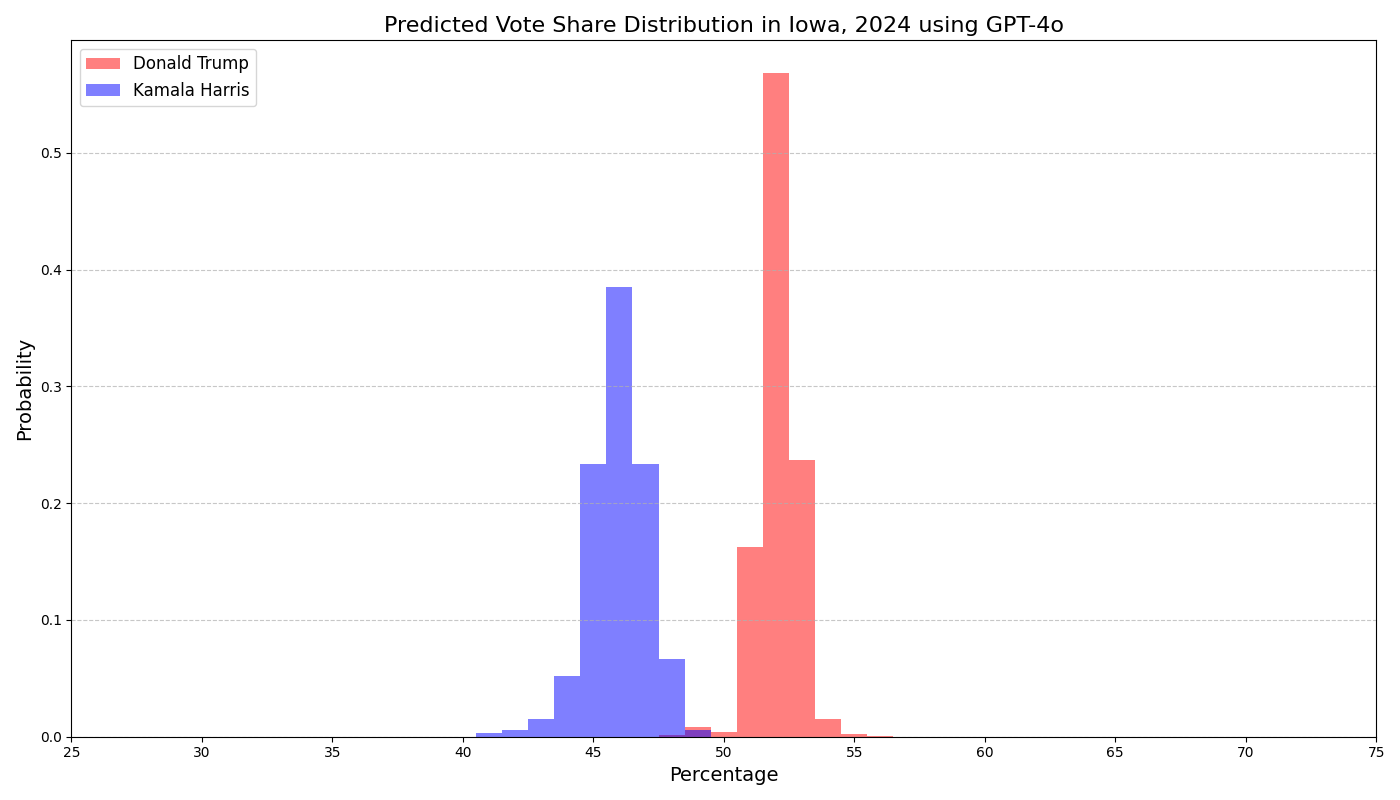}
    \caption{Model predictions for Iowa (typically a Republican leaning state) in the 2024 presidential election, showing a distribution for the model's prediction that is distinct from its prediction for the 2020 election.}
    \label{fig:model_predicted_2020}
\end{figure}

\begin{figure}[H]
    \centering
    \includegraphics[width=0.6\textwidth]{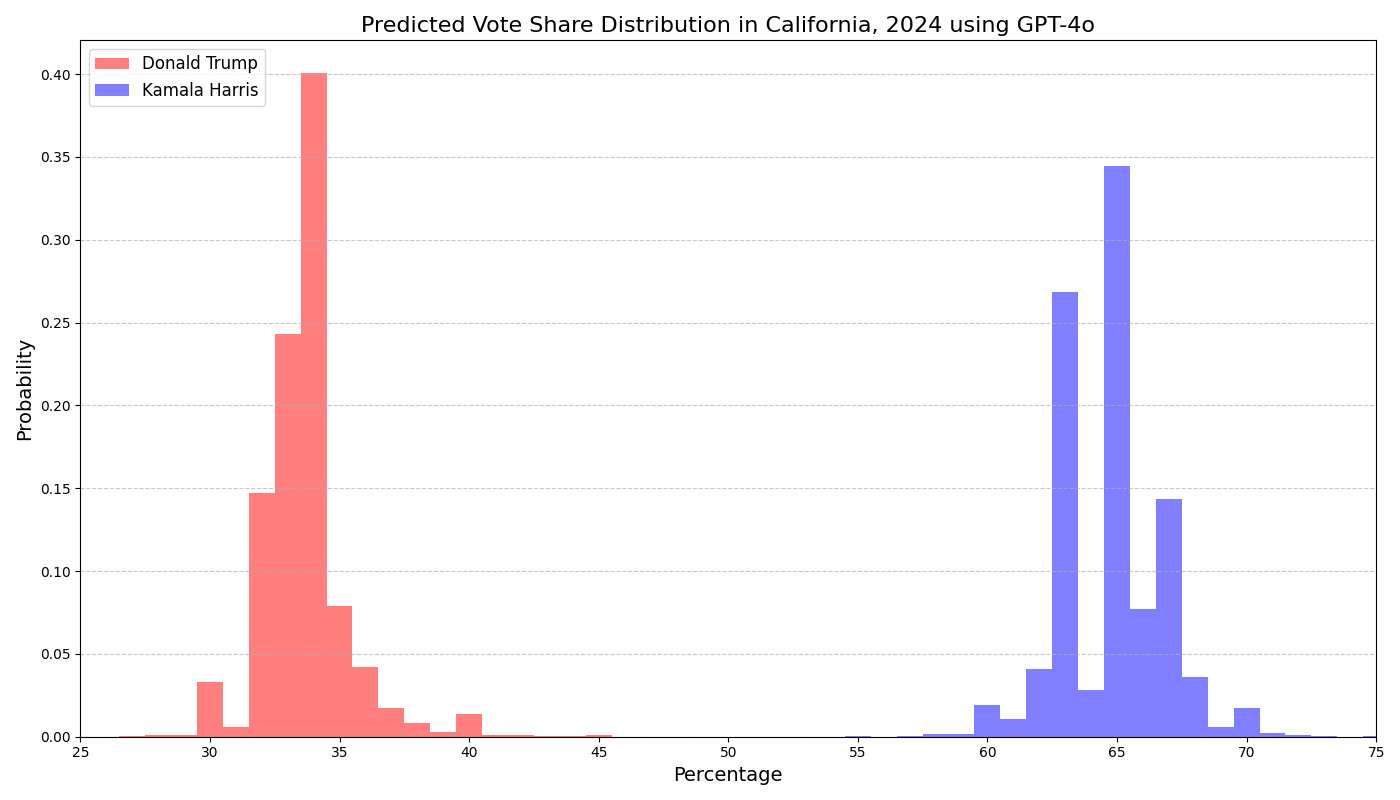}
    \caption{Model predictions for California (a strongly Democratic state)}
    \label{fig:model_predicted_2020}
\end{figure}

\begin{figure}[H]
    \centering
    \includegraphics[width=0.6\textwidth]{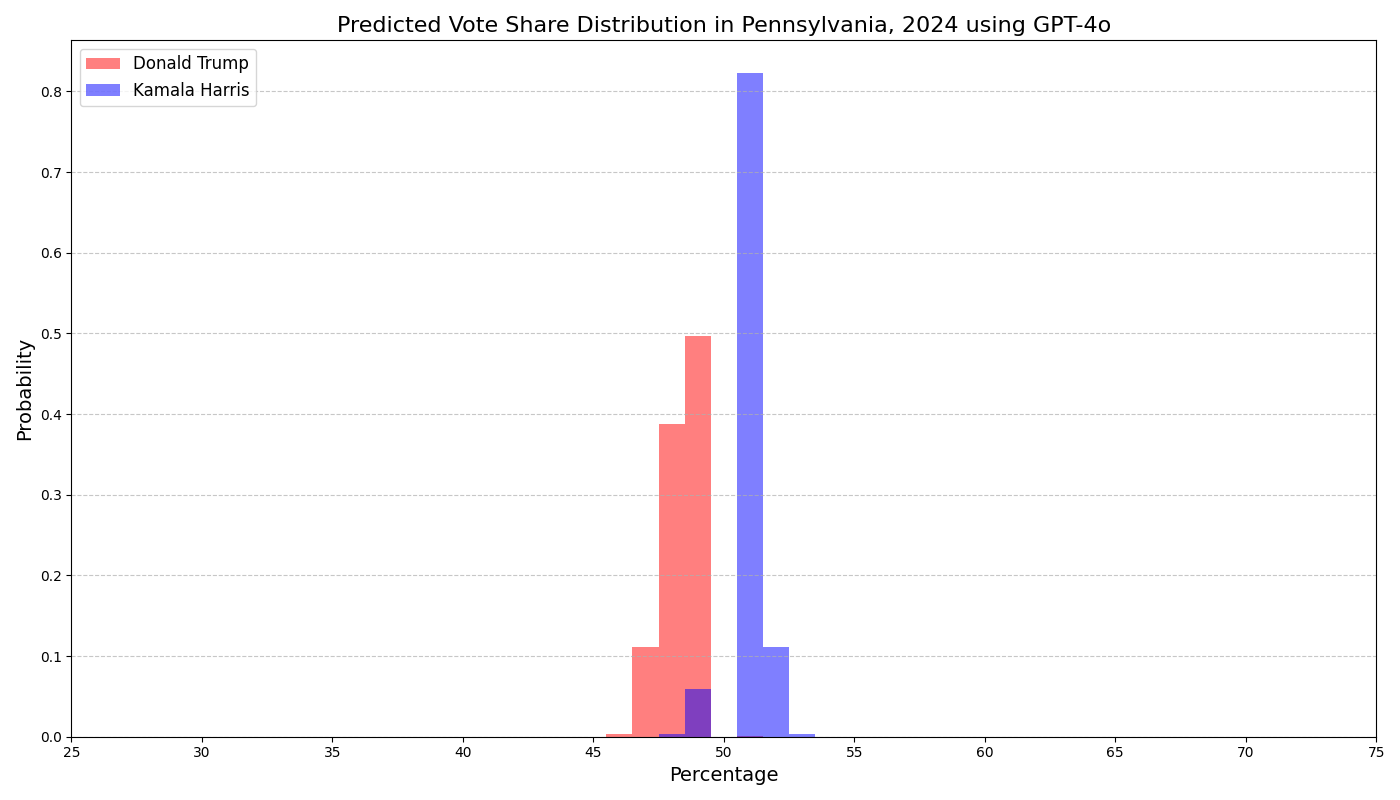}
    \caption{Model predictions for Pennsylvania (a swing state)}
    \label{fig:model_predicted_2020}
\end{figure}

\subsection{Predicted Electoral Outcomes}
Using these predicted state vote distributions, we run our electoral college simulation for each contest and calculate the result of a simulated presidential election between Donald Trump and Kamala Harris. Our output is a probability distribution over all possible electoral college outcomes. To produce a single electoral map, however, we simply pit the weighted average of each candidate-state distribution against their opponent to produce an average case result.

\begin{figure}[H]
    \centering
    \includegraphics[width=0.6\textwidth]{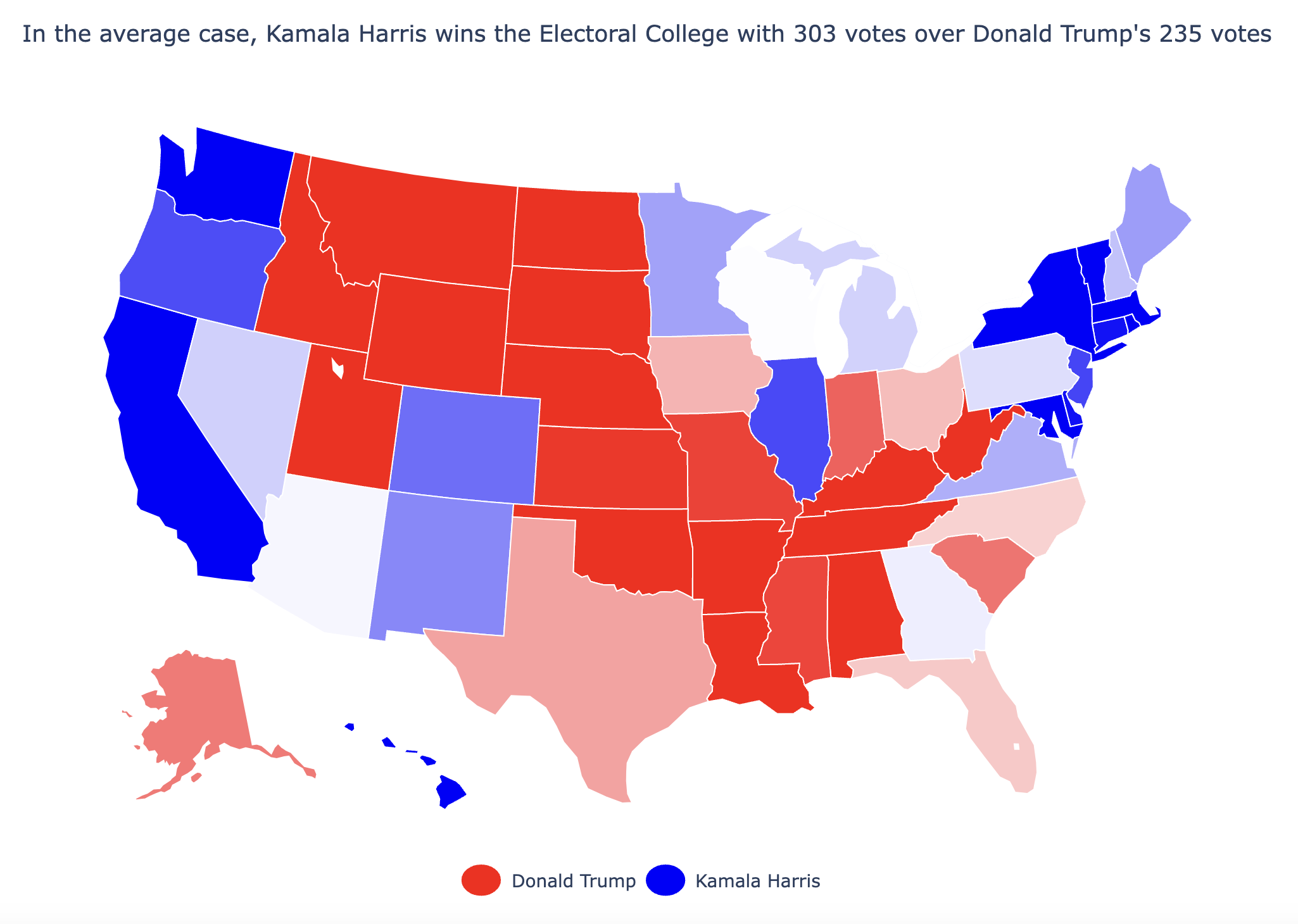}
    \caption{Predicted result of the electoral college of the 2024 presidential election. Kamala Harris is predicted to win the electoral college with 303 votes over Donald Trump's 235 votes. From the electoral probability distribution, the model predicts that Kamala Harris wins in 95\% of all cases.}
    \label{fig:model_predicted_2020}
\end{figure}

\subsection{Simulating Hypothetical Elections}
Using this method of generating electoral results, it's trivial to change the input names and specified election to simulate election results for any two individuals in a given contest. We simulate hypothetical match-ups between five Republican and five Democratic politicians in the 2016, 2020, and 2024 presidential elections. Even though the first two elections have already occurred and the results figure in the LLM's training data, we still phrase the prompt as if it were a prediction.

\begin{figure}[H]
    \centering
    \begin{subfigure}[b]{0.32\textwidth}
        \centering
        \includegraphics[width=\textwidth]{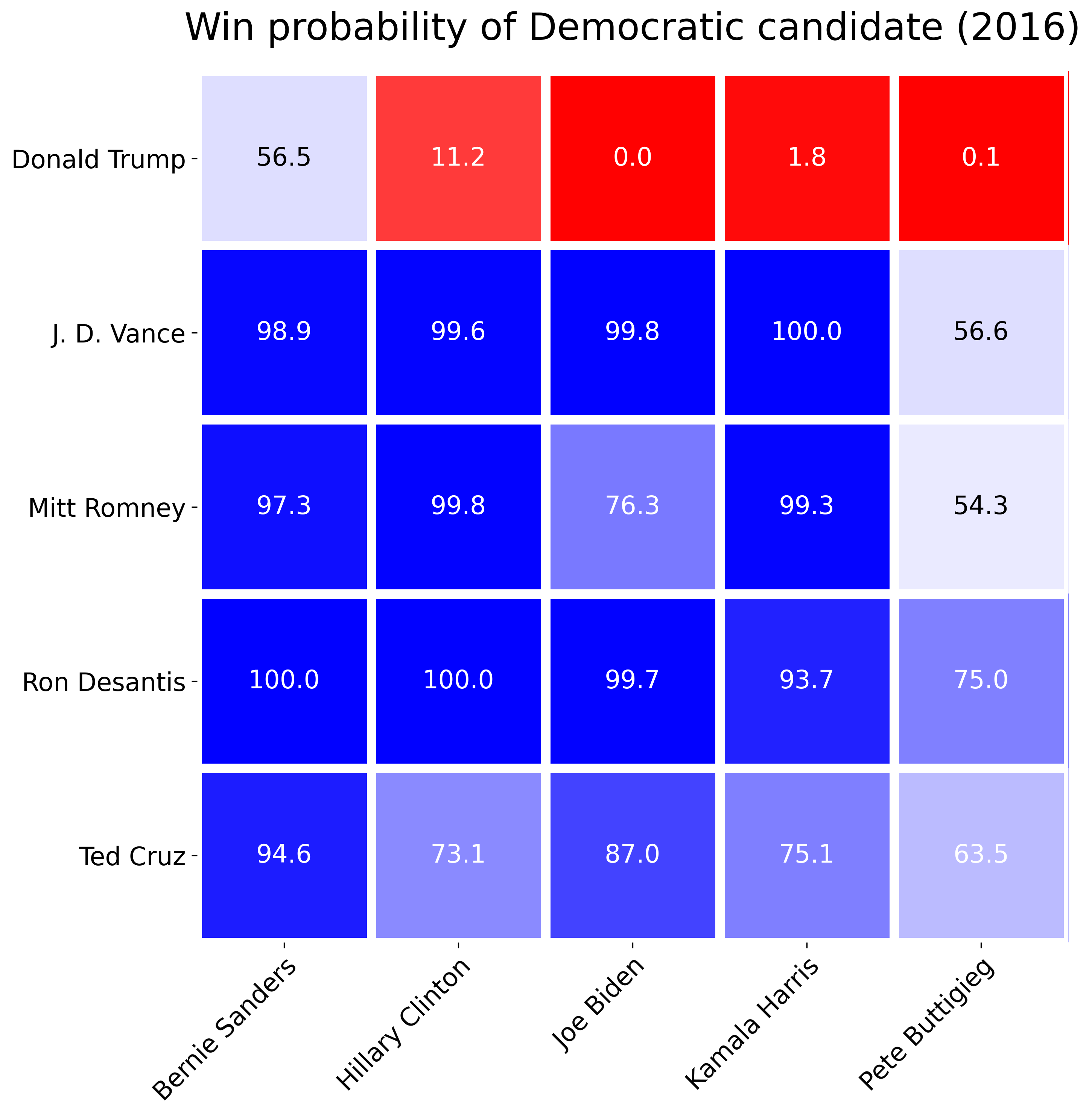}
        \caption{2016 matchups.}
        \label{fig:model_predicted_2016}
    \end{subfigure}
    \hfill
    \begin{subfigure}[b]{0.32\textwidth}
        \centering
        \includegraphics[width=\textwidth]{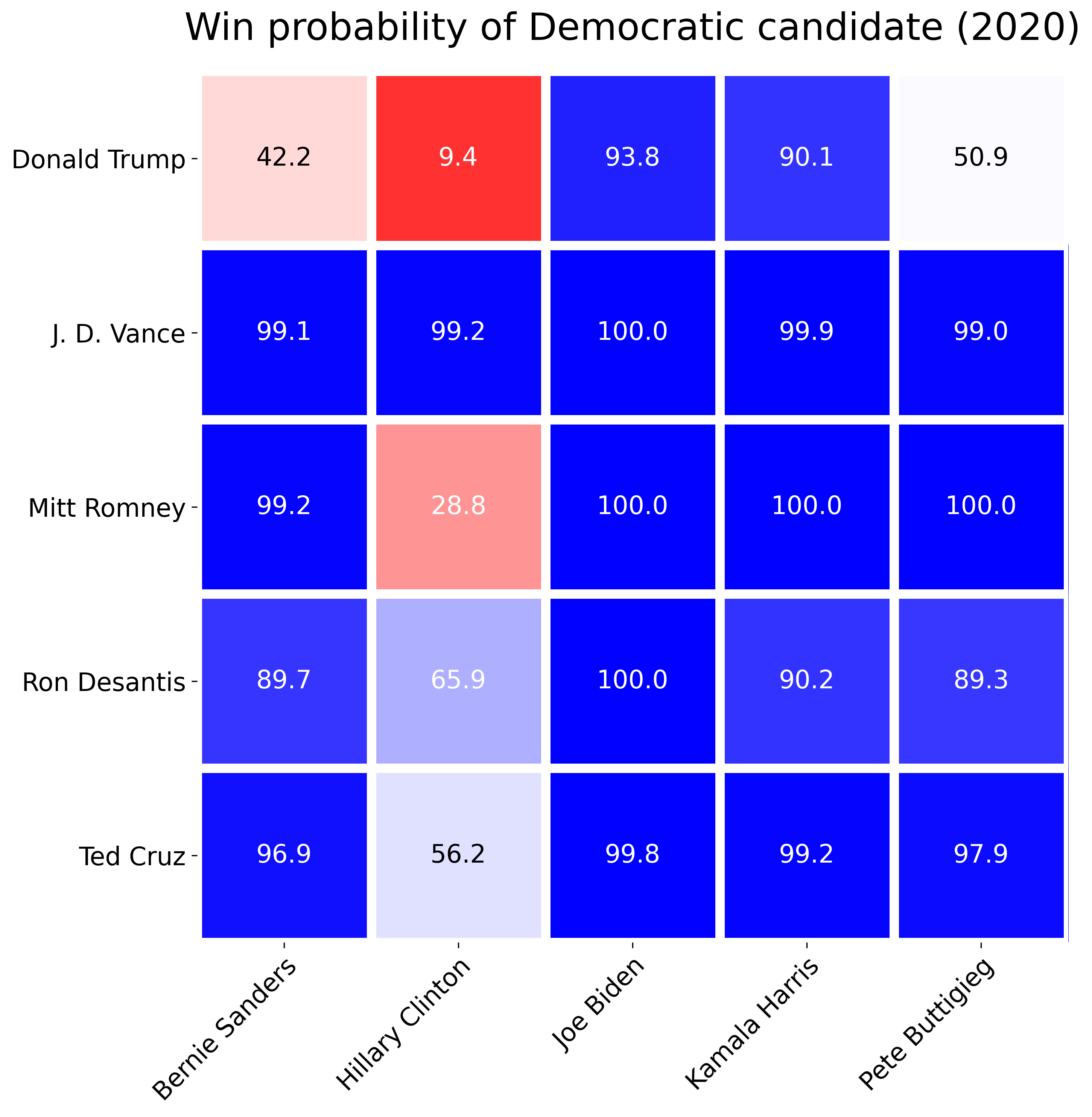}
        \caption{2020 matchups.}
        \label{fig:model_predicted_2020}
    \end{subfigure}
    \hfill
    \begin{subfigure}[b]{0.32\textwidth}
        \centering
        \includegraphics[width=\textwidth]{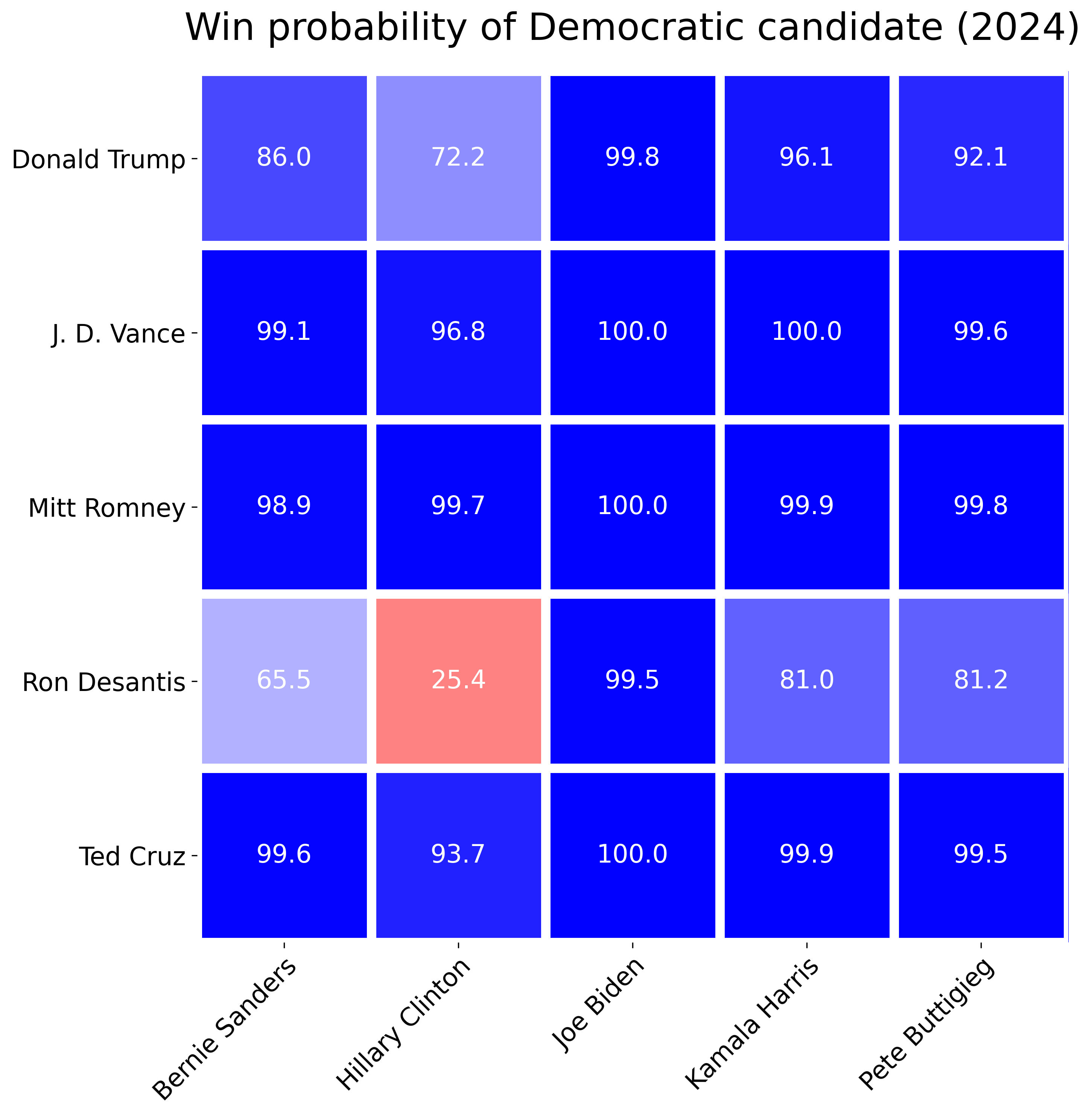}
        \caption{2024 matchups.}
        \label{fig:model_predicted_2024}
    \end{subfigure}
    \caption{Comparison of predicted electoral college outcomes for hypothetical matchups in the 2016, 2020, and 2024 elections. Bluer shades indicate a higher probability of a Democratic win, while red shades indicate a Republican advantage. Note that the model correctly recalled Donald Trump's 2016 win, but otherwise strongly favors the Democratic Party.}
    \label{fig:matchup_comparison}
\end{figure}

\section{Discussion}
\subsection{Prompt Noise}
One of the major challenges with using LLMs is the impact that prompt noise can have on generated outputs, to the extent that using different wordings can significantly alter the output distribution \cite{sun2024investigationpromptvariationszeroshot}. Our proposed approach is not exempt from this issue, but it may provide an additional tool set for analyzing the impact of prompt noise on applicable tasks by comparing output distributions across a variety of similar prompts. 

\subsection{Measuring Bias}
Using direct prediction, extracted distributions can be compared to actual results as a means of assessing an LLMs world model bias for a given task. This would be most effective on tasks with more data points, as it would ideally reduce the impact of outliers. This is one of the key motivations for our simulating the election starting at the state rather than national level, as it provides 102 data points rather than 2. 

Given the context of United States politics, several major LLMs have been shown to have left leaning biases \cite{potter2024hiddenpersuadersllmspolitical}. Initial testing with direct prediction has reflected this biased world model, with LLMs generally favoring Democratic candidates over Republican candidates in swing states, where the vote could go to either party. Notably, this bias does not seem to impact non-swing states, where a given party is assumed to have an easy victory. As shown in Figure \ref{fig:matchup_comparison}, the LLM's predictions consistently favor Democratic candidates, except in the case of Donald Trump’s dominance in the 2016 election predictions. This exception likely reflects the influence of his victory, which is included in the model’s training data.

\subsection{Limitations of Training Data Cutoff}
Retraining or repeatedly fine tuning an LLM is computationally expensive and runs the risk of issues such as catastrophic forgetting. Furthermore, the data must be carefully filtered to limit harmful content or misinformation. As such, an LLM's training data is generally cut off significantly before its release, preventing its world model from accounting for current events. For example, the cut off for GPT-4o was October 2023, 7 months before its May 2024 release.

The two most straightforward techniques for overcoming this limitation are supplementing the prompt or fine tuning with current information. However, both of these approaches introduce additional challenge, such as determining what information should be included in a summary of current events. 

This limitation also provides opportunity for the study of an LLM as a predictive model, as we can compare output distributions of current events not seen by the model to the actual events.

\subsection{Measure of Algorithmic Fidelity For LLMs as Predictive Models}
By extracting distributions that can be compared to actual data, we obtain a new method of determining a models algorithmic fidelity. Although limited to a case by case basis (fidelity to voting trends does not inform us about a model's ability to predict sports games), measuring algorithmic fidelity in this way can provide a measure of an LLM's effectiveness as a predictive model in a given domain. To what extent this approach to measuring algorithmic fidelity is valid or consistent is yet to be seen.

\section{Conclusion}
In this paper we propose Distribution Based Prediction, a technique that involves treating an LLM's output probabilities as a distribution representative of the LLM's internal world model. This method allows for a comparison against ground truth data as a means of assessing model bias and algorithmic fidelity.

Initial testing of this approach for simulating the US presidential election indicates that Distribution Based Prediction could be a viable method of utilizing and analyzing LLMs as predictive models.

\section{Future Work}
A great deal of work is needed to verify the effectiveness of distribution based prediction as a framework. This includes determining this method's robustness against prompt noise, creating a rigorous measure of algorithmic fidelity in this context, and testing across additional domains.

One of the key aspects of distribution based prediction is its generalization - nearly any task where the data can be expressed as a two dimensional distribution may be modeled. Use cases could include predicting the results of sports games, forecasting the weather, predicting stock market trends, and much more. Additional research into any of these topics using this approach could be valuable, both to the study of LLMs and the specific domain.

As far as US presidential election electoral prediction, additional work is needed to probe model bias, compare model outputs to real data, and measure the effect of adding more up to date information. In particular, the actual results in each state may be compared to an aggregation of model outputs across many prompts, allowing us to determine how likely it is that the real distribution that reality drew from matches the LLM's world model.

\bibliography{LLMVP}

\begin{thebibliography}{10}

\bibitem{Argyle_2023}
Lisa~P. Argyle, Ethan~C. Busby, Nancy Fulda, Joshua~R. Gubler, Christopher Rytting, and David Wingate.
\newblock Out of one, many: Using language models to simulate human samples.
\newblock {\em Political Analysis}, 31(3):337–351, February 2023.

\bibitem{Bisbee_Clinton_Dorff_Kenkel_Larson_2024}
James Bisbee, Joshua~D. Clinton, Cassy Dorff, Brenton Kenkel, and Jennifer~M. Larson.
\newblock Synthetic replacements for human survey data? the perils of large language models.
\newblock {\em Political Analysis}, 32(4):401–416, 2024.

\bibitem{borisov2023languagemodelsrealistictabular}
Vadim Borisov, Kathrin Seßler, Tobias Leemann, Martin Pawelczyk, and Gjergji Kasneci.
\newblock Language models are realistic tabular data generators, 2023.

\bibitem{feng2023pretrainingdatalanguagemodels}
Shangbin Feng, Chan~Young Park, Yuhan Liu, and Yulia Tsvetkov.
\newblock From pretraining data to language models to downstream tasks: Tracking the trails of political biases leading to unfair nlp models, 2023.

\bibitem{gujral2024llmshelppredictelections}
Pratik Gujral, Kshitij Awaldhi, Navya Jain, Bhavuk Bhandula, and Abhijnan Chakraborty.
\newblock Can llms help predict elections? (counter)evidence from the world's largest democracy, 2024.

\bibitem{gurnee2024languagemodelsrepresentspace}
Wes Gurnee and Max Tegmark.
\newblock Language models represent space and time, 2024.

\bibitem{li2024emergentworldrepresentationsexploring}
Kenneth Li, Aspen~K. Hopkins, David Bau, Fernanda Viégas, Hanspeter Pfister, and Martin Wattenberg.
\newblock Emergent world representations: Exploring a sequence model trained on a synthetic task, 2024.

\bibitem{liu2024largelanguagemodelsassume}
Ryan Liu, Jiayi Geng, Joshua~C. Peterson, Ilia Sucholutsky, and Thomas~L. Griffiths.
\newblock Large language models assume people are more rational than we really are, 2024.

\bibitem{ong2024gptologycomputationalmodelssilicon}
Desmond~C. Ong.
\newblock Gpt-ology, computational models, silicon sampling: How should we think about llms in cognitive science?, 2024.

\bibitem{potter2024hiddenpersuadersllmspolitical}
Yujin Potter, Shiyang Lai, Junsol Kim, James Evans, and Dawn Song.
\newblock Hidden persuaders: Llms' political leaning and their influence on voters, 2024.

\bibitem{rozado2024politicalpreferencesllms}
David Rozado.
\newblock The political preferences of llms, 2024.

\bibitem{marketingsilsample2024}
Marko Sarstedt, Susanne Adler, Lea Rau, and Bernd Schmitt.
\newblock Using large language models to generate silicon samples in consumer and marketing research: Challenges, opportunities, and guidelines.
\newblock {\em Psychology and Marketing}, 41:n/a--n/a, 02 2024.

\bibitem{sun2024randomsiliconsamplingsimulating}
Seungjong Sun, Eungu Lee, Dongyan Nan, Xiangying Zhao, Wonbyung Lee, Bernard~J. Jansen, and Jang~Hyun Kim.
\newblock Random silicon sampling: Simulating human sub-population opinion using a large language model based on group-level demographic information, 2024.

\bibitem{sun2024investigationpromptvariationszeroshot}
Shuoqi Sun, Shengyao Zhuang, Shuai Wang, and Guido Zuccon.
\newblock An investigation of prompt variations for zero-shot llm-based rankers, 2024.

\bibitem{RePEc:osf:socarx:97r8s}
Leah von~der Heyde, Anna-Carolina Haensch, and Alexander Wenz.
\newblock {Assessing Bias in LLM-Generated Synthetic Datasets: The Case of German Voter Behavior}.
\newblock SocArXiv 97r8s, Center for Open Science, December 2023.

\bibitem{wei2022emergentabilitieslargelanguage}
Jason Wei, Yi~Tay, Rishi Bommasani, Colin Raffel, Barret Zoph, Sebastian Borgeaud, Dani Yogatama, Maarten Bosma, Denny Zhou, Donald Metzler, Ed~H. Chi, Tatsunori Hashimoto, Oriol Vinyals, Percy Liang, Jeff Dean, and William Fedus.
\newblock Emergent abilities of large language models, 2022.

\bibitem{yang2024llmvotinghumanchoices}
Joshua~C. Yang, Damian Dailisan, Marcin Korecki, Carina~I. Hausladen, and Dirk Helbing.
\newblock Llm voting: Human choices and ai collective decision making, 2024.

\bibitem{yang2024largelanguagemodelsllms}
Kaiqi Yang, Hang Li, Hongzhi Wen, Tai-Quan Peng, Jiliang Tang, and Hui Liu.
\newblock Are large language models (llms) good social predictors?, 2024.

\end{thebibliography}
\bibliographystyle{plain}

\appendix

\section{Impact of Model Size on Predicted Voter Distributions}
\label{appendix:model_size}

\begin{figure}[H]    
\centering
\begin{subfigure}[t]{0.48\textwidth}
    \centering
    \includegraphics[width=\textwidth]{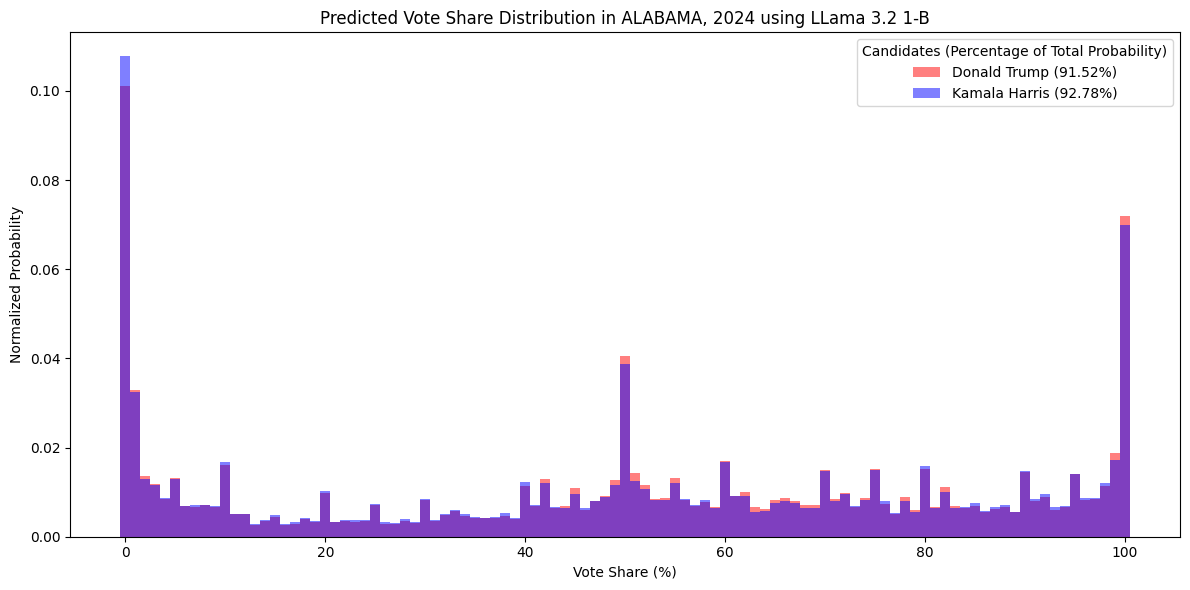}
    \caption{Alabama - 1B Parameters}
\end{subfigure}%
\hfill
\begin{subfigure}[t]{0.48\textwidth}
    \centering
    \includegraphics[width=\textwidth]{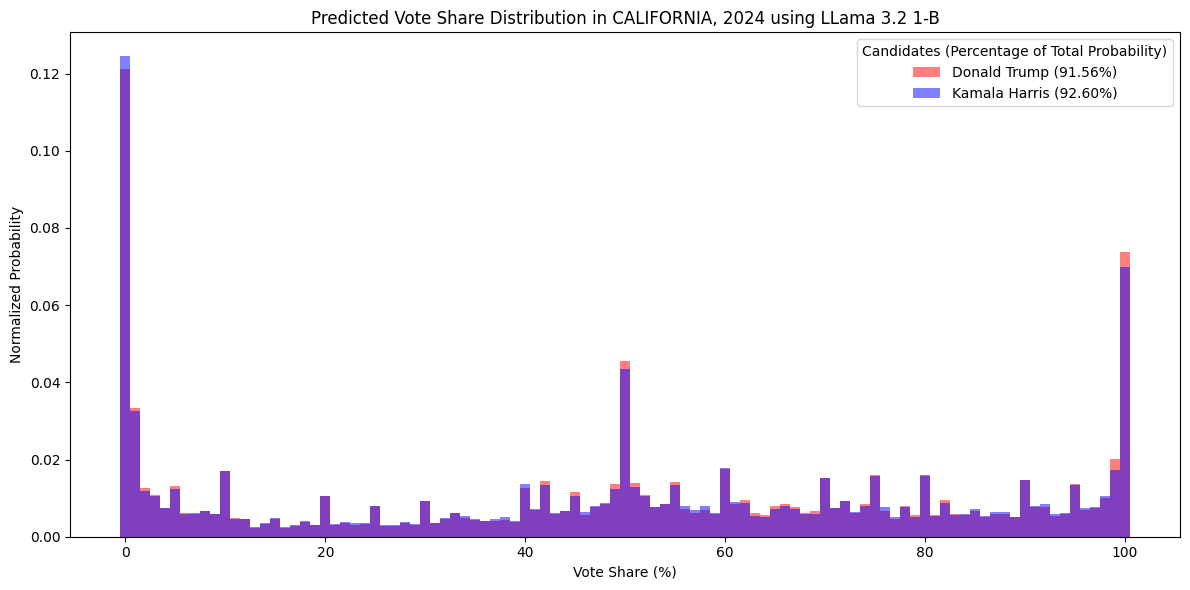}
    \caption{California - 1B Parameters}
\end{subfigure}

\vspace{0.3cm}

\begin{subfigure}[t]{0.48\textwidth}
    \centering
    \includegraphics[width=\textwidth]{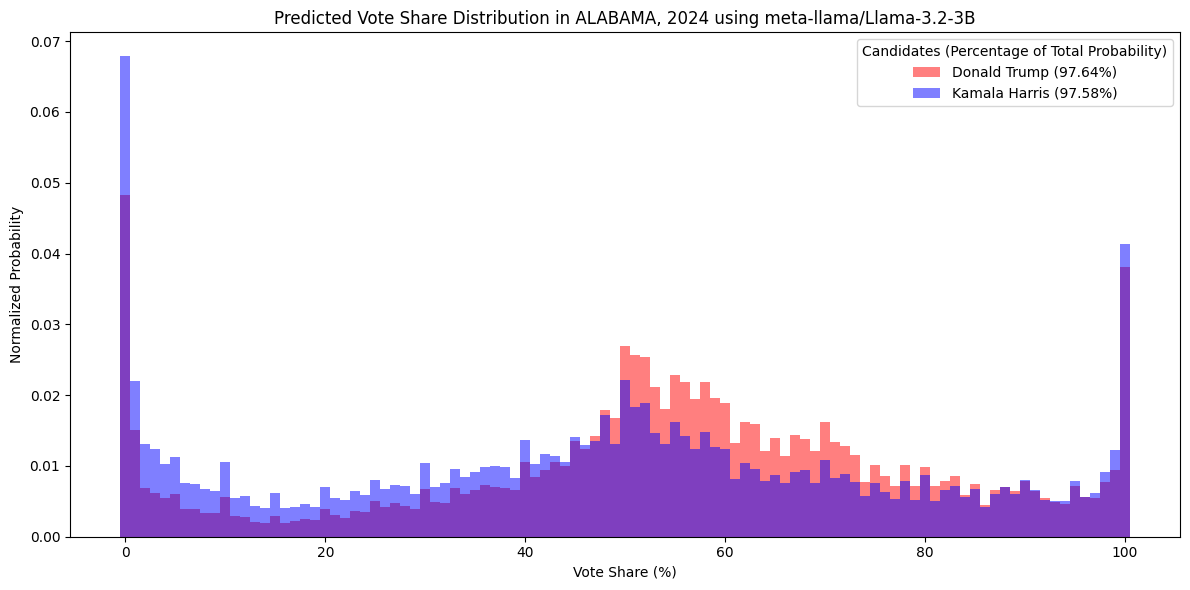}
    \caption{Alabama - 3B Parameters}
\end{subfigure}%
\hfill
\begin{subfigure}[t]{0.48\textwidth}
    \centering
    \includegraphics[width=\textwidth]{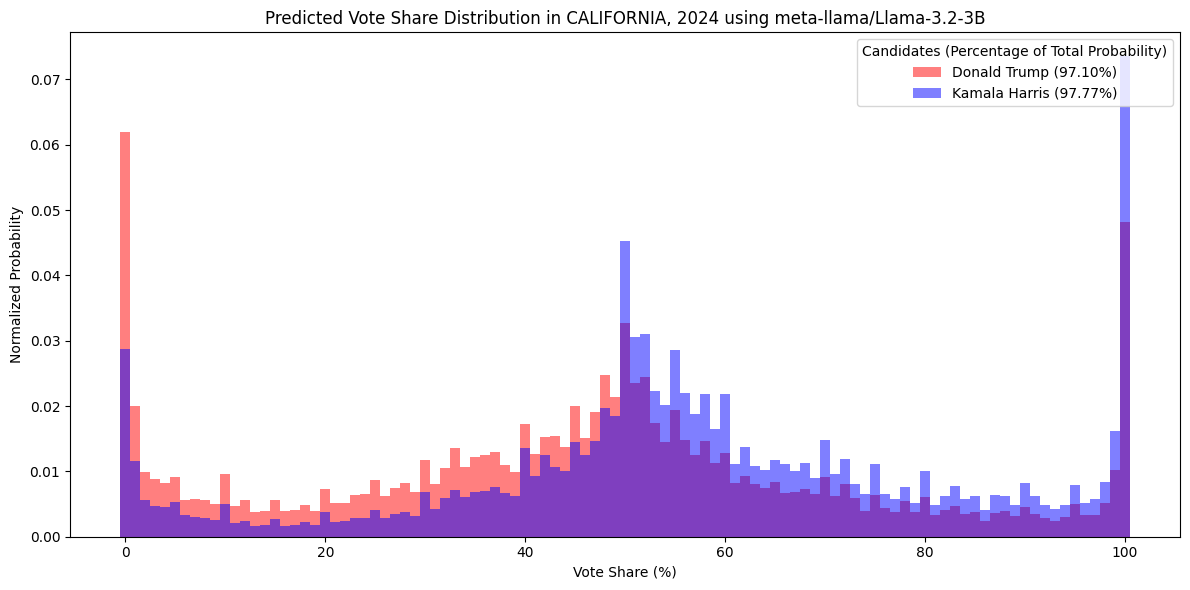}
    \caption{California - 3B Parameters}
\end{subfigure}

\vspace{0.3cm}

\begin{subfigure}[t]{0.48\textwidth}
    \centering
    \includegraphics[width=\textwidth]{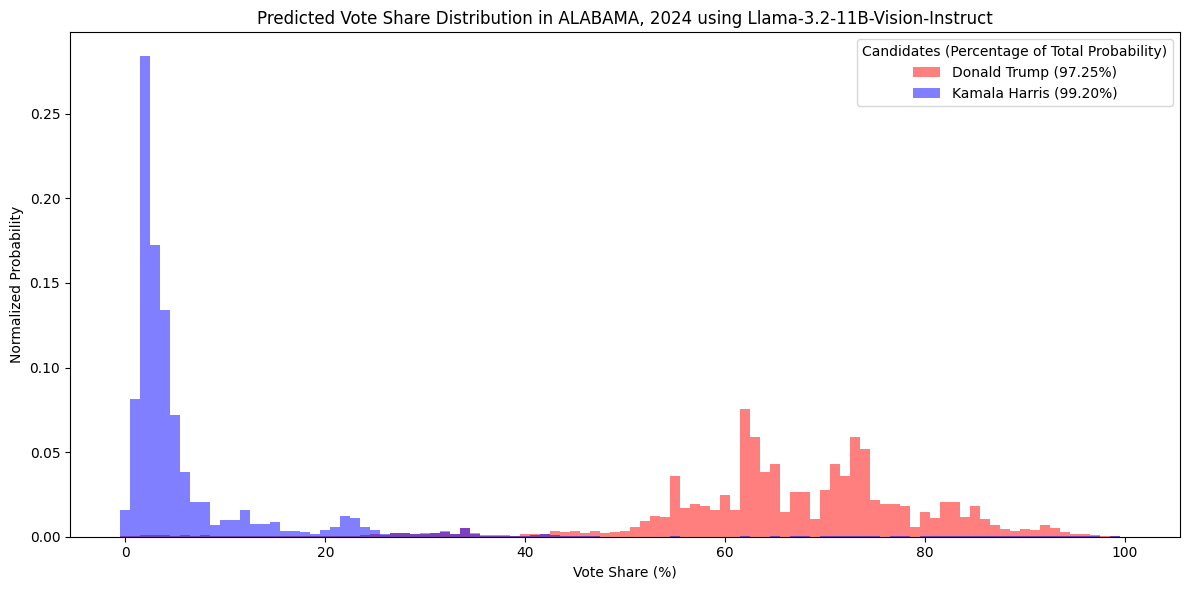}
    \caption{Alabama - 11B Parameters}
\end{subfigure}%
\hfill
\begin{subfigure}[t]{0.48\textwidth}
    \centering
    \includegraphics[width=\textwidth]{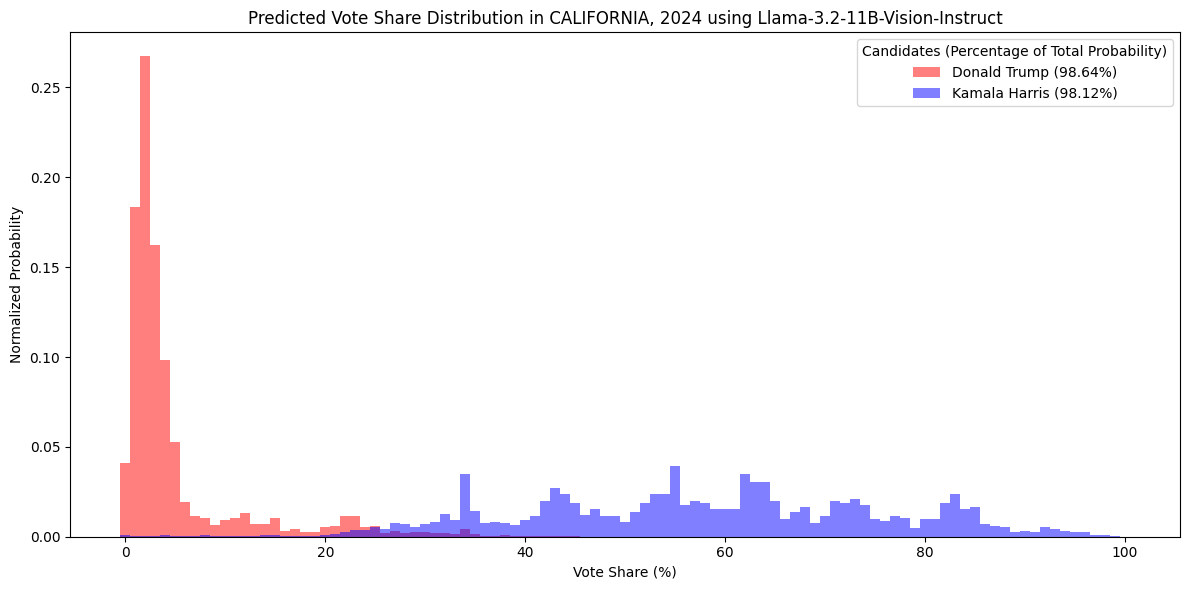}
    \caption{California - 11B Parameters}
\end{subfigure}

\vspace{0.3cm}

\begin{subfigure}[t]{0.48\textwidth}
    \centering
    \includegraphics[width=\textwidth]{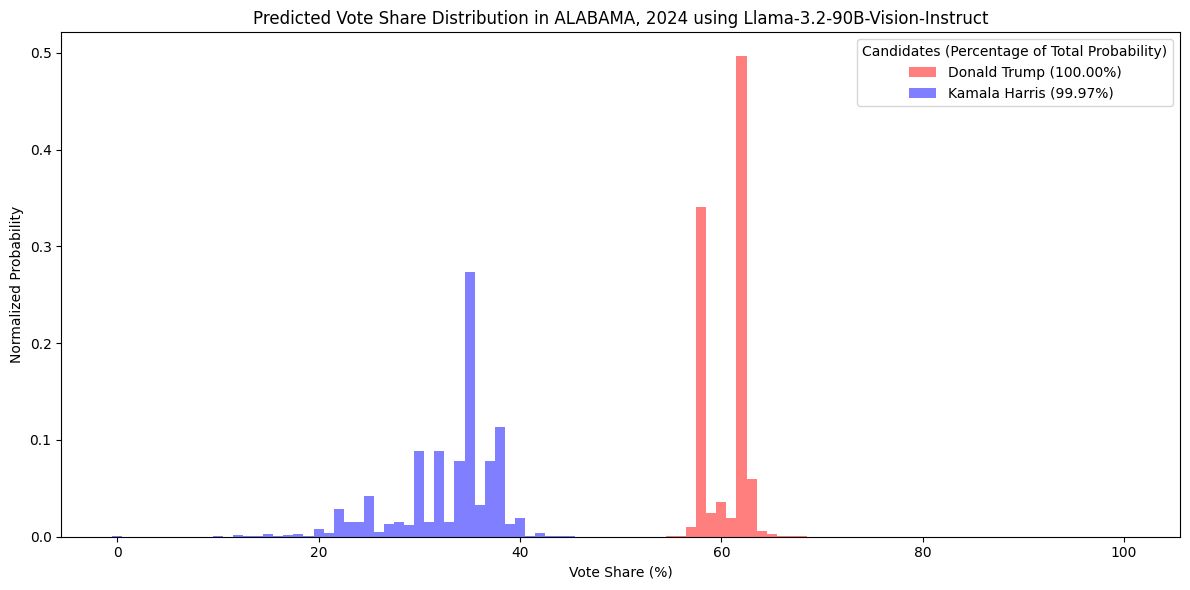}
    \caption{Alabama - 90B Parameters}
\end{subfigure}%
\hfill
\begin{subfigure}[t]{0.48\textwidth}
    \centering
    \includegraphics[width=\textwidth]{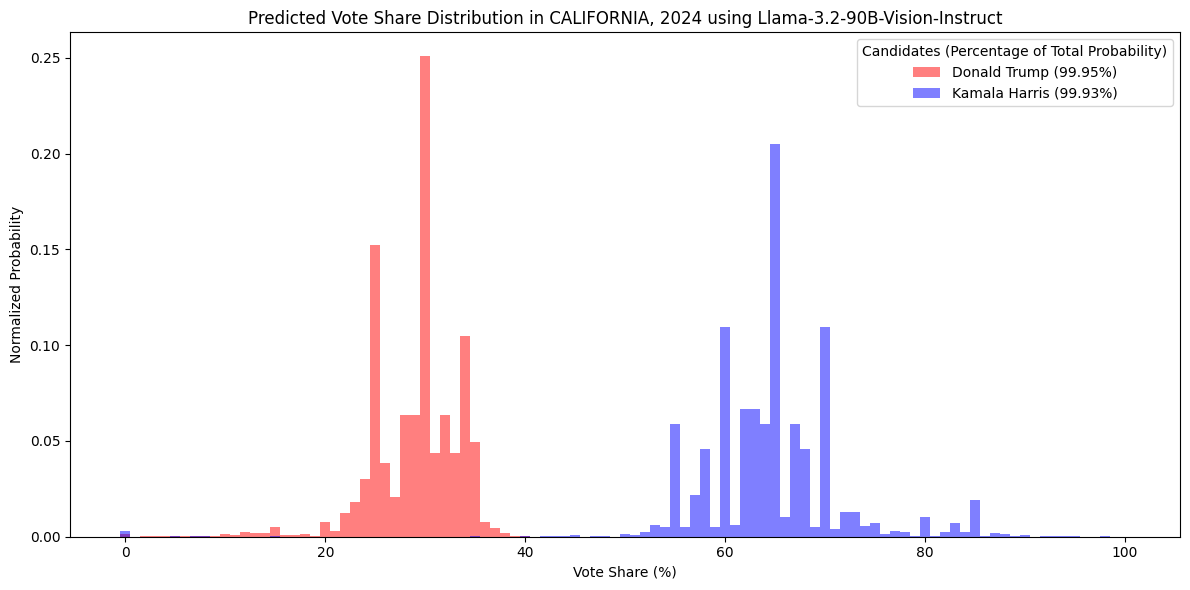}
    \caption{California - 90B Parameters}
\end{subfigure}

\caption{Comparison of distributions of predicted election outcomes  (measured in percent of vote share) for Donald Trump and Kamala Harris in simulated elections held in Alabama and California. Results are plotted for 4 different sizes of the Llama 3.2 model. Note that California is pro Democratic Party (blue) and Alabama is pro Republican Party (red). Probability associated with candidates describes what percent of the predicted tokens conformed with the expected output format (integers). }
\label{fig:model_size}
\end{figure}

We input the same system and user prompts to Llama 3.2 models of increasing sizes and observe how the output distributions differ. 

\begin{itemize}
\item \textbf{System Prompt:} \\
\textit{``You are an impartial election prediction machine. Respond with a single integer token between 0 and 100 representing the vote share."}

\item \textbf{User Prompt:} \\
\textit{``What percentage of the vote will \{candidate\} win in the \{year\} presidential election in \{state\}?"}
\end{itemize}

Figure~\ref{fig:model_size} illustrates how increasing the size of the model allows for better world models, allowing more fidelity to real trends.  The smallest model, with 1 billion parameters, understands that the prompt requires an output integer but does not significantly differentiate between the states or candidates. The 3 billion parameter model begins to demonstrate a more nuanced understanding, slightly favoring the Republican party in Alabama and the Democratic party in California. At 11 billion parameters, the model strongly favors the correct parties in each state, almost entirely differentiating between the two parties distributions. Despite this, the distributions are illogical, in that combining them rarely accounts for most of the states voters. Finally, the 90 billion parameter model clearly differentiates the parties into logical distributions that seem to compliment each other, at least appearing similar to actual voter distributions.

\section{Tendency of Models to Stereotype During Silicon Sampling}
\label{appendix:silicon_sampling}
Initially, we explored a demographic-based simulation approach, aiming to leverage the model’s embedded knowledge about voter demographics across geographic regions. The idea was to simulate a diverse population of “voters” through Silicon Sampling, prompting the model to generate responses based on specific demographic information (e.g., rural Christians in Pennsylvania). This approach, however, revealed limitations: the LLM frequently displayed strong stereotypical biases, repeatedly associating certain demographic groups with uniform voting behaviors. For instance, when prompted to “sample” the preferences of a rural Christian Pennsylvanian, the model overwhelmingly leaned toward the voting patterns of the majority demographic preference, with minimal variance in outcomes.

To illustrate this process, we used the following structured prompts for guiding the model’s responses in the demographic sampling approach:

\begin{itemize}
\item \textbf{System Prompt:} \\
\textit{``You are a \{demographic\} American. Not the average of all \{demographic\} Americans. You are literally a random \{demographic\} American sampled from the population. You respond by picking between two candidates for the presidential election of a given year by outputting only the last name of your chosen candidate."}

\item \textbf{User Prompt:} \\
\textit{``Kamala Harris (the Democratic candidate) is running against Donald Trump (the Republican candidate) in the 2024 presidential election. You cast your vote for:"}
\end{itemize}

\begin{figure}[htbp]
    \centering
    \hspace*{-1cm}\includegraphics[width=1.2\textwidth]{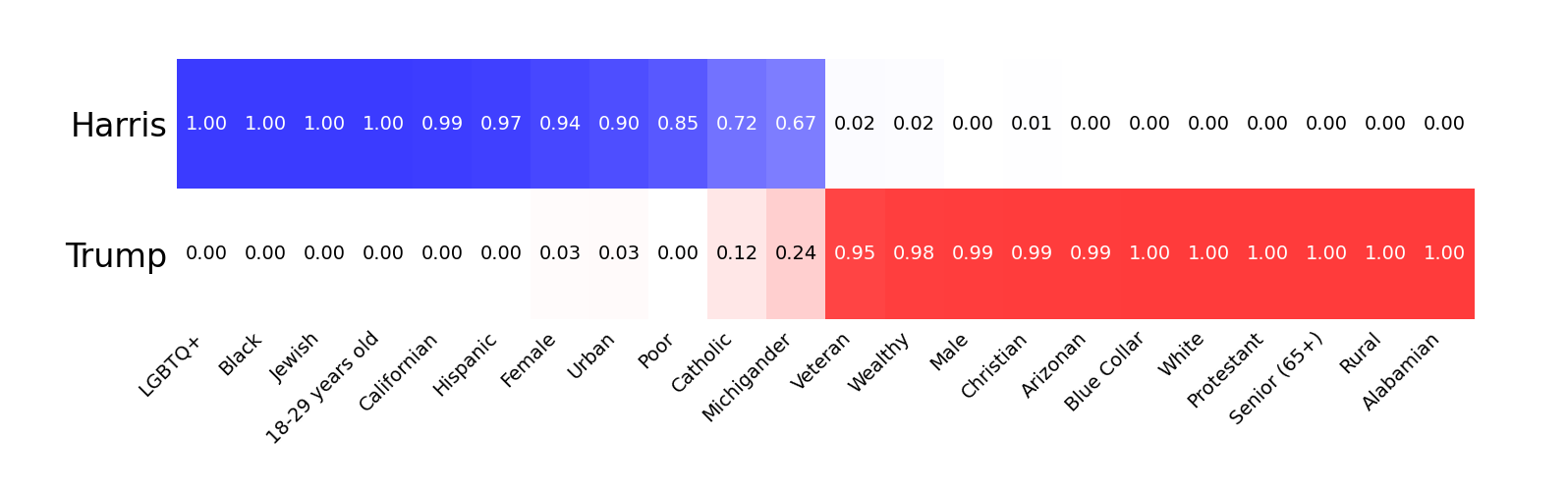}
    \caption{Token probabilities across different demographics. Note that the model tends to hallucinate a probability for the token ``Biden" for various demographics such as ``Catholic" or ``Michigander" which results in the probabilities adding up to less than 1.}
    \label{fig:limited_demographic_probabilities}
\end{figure}

This approach, while conceptually promising for simulating demographic samples, underscored the limitations of LLMs in representing the gradations within demographic groups through Silicon Sampling. As illustrated in Figure \ref{fig:limited_demographic_probabilities}, token probabilities almost entirely align with the stereotypical majority-class outcomes, producing skewed distributions that failed to reflect the true diversity of political behavior within and across demographic segments. Further reinforcing this concern, Bisbee et. al. show this approach not only has less variability than real surveys, but is also highly susceptible to prompt noise \cite{Bisbee_Clinton_Dorff_Kenkel_Larson_2024}.

\begin{table}[h!]
\centering
\begin{tabular}{|l|c|c|}
\hline
\textbf{State} & \textbf{Joe Biden Error} & \textbf{Donald Trump Error} \\
\hline
Alaska & 0.0005 & 0.1003 \\
Hawaii & 0.0502 & 0.3476 \\
Washington & 0.4001 & 0.9775 \\
Oregon & 0.0480 & 0.6887 \\
California & 0.4148 & 0.3000 \\
Idaho & 0.0737 & 0.1000 \\
Montana & 0.3820 & 0.0999 \\
Nevada & 0.9000 & 0.6997 \\
Wyoming & 0.2941 & 0.4041 \\
Utah & 0.0563 & 0.1860 \\
Arizona & 0.5994 & 0.1000 \\
Colorado & 0.1982 & 0.1209 \\
New Mexico & 0.2994 & 0.3355 \\
North Dakota & 0.0978 & 0.4969 \\
South Dakota & 0.5113 & 0.2000 \\
Nebraska & 0.2400 & 0.4325 \\
Kansas & 0.0308 & 0.4859 \\
Oklahoma & 0.6266 & 0.3983 \\
Texas & 0.4797 & 0.0981 \\
Minnesota & 0.5981 & 0.9190 \\
Iowa & 0.0097 & 0.1999 \\
Missouri & 0.3966 & 0.1998 \\
Arkansas & 0.1742 & 0.3627 \\
Louisiana & 0.1000 & 0.3933 \\
Wisconsin & 0.5975 & 0.0859 \\
Illinois & 0.5993 & 0.6676 \\
Mississippi & 0.0139 & 0.4355 \\
Michigan & 0.2933 & 0.1878 \\
Indiana & 0.0022 & 0.0988 \\
Kentucky & 3.0925 & 0.0933 \\
Tennessee & 0.3984 & 0.2860 \\
Alabama & 0.2931 & 0.1998 \\
Ohio & 0.0205 & 0.3000 \\
West Virginia & 0.2988 & 0.3978 \\
Georgia & 0.4988 & 0.3000 \\
Florida & 0.0998 & 0.2000 \\
Maine & 0.1002 & 0.0223 \\
New Hampshire & 0.0998 & 0.5262 \\
Vermont & 0.3990 & 0.2370 \\
Massachusetts & 0.0949 & 0.6151 \\
Rhode Island & 0.6772 & 0.2010 \\
Connecticut & 0.4779 & 0.1497 \\
New York & 3.8997 & 4.1224 \\
Pennsylvania & 1.0002 & 0.1237 \\
Virginia & 0.3996 & 0.1997 \\
New Jersey & 0.2841 & 0.3998 \\
Delaware & 0.1865 & 0.3815 \\
Maryland & 0.7797 & 0.3610 \\
North Carolina & 0.2972 & 1.0993 \\
South Carolina & 0.3630 & 0.0964 \\
District Of Columbia & 0.6521 & 1.7713 \\
\textbf{Average Error} & \textbf{0.4490} & \textbf{0.4354} \\
\textbf{Standard Deviation} & \textbf{0.6734} & \textbf{0.6120} \\
\hline
\end{tabular}
\caption{State-by-state error between ground truth and weighted average of generated distributions for Joe Biden and Donald Trump in the 2020 presidential election.}
\label{tab:state_differences}
\end{table}

\end{document}